\documentclass[letterpaper, 10 pt, conference]{ieeeconf}  

\IEEEoverridecommandlockouts                              

\usepackage[accsupp]{axessibility}
\usepackage{graphicx}
\usepackage{amsmath}
\usepackage{amssymb}
\usepackage{booktabs}
\usepackage{mathtools}
\usepackage[x11names, table]{xcolor}
\usepackage{scalerel}
\usepackage{balance}
\usepackage{hhline}
\usepackage{multirow}
\usepackage{bm}
\usepackage{placeins} 
\usepackage{flafter}  
\usepackage{lipsum}

\DeclareMathOperator*{\argmin}{\arg\!\min}

\usepackage[ruled,linesnumbered]{algorithm2e}

\makeatletter
\let\NAT@parse\undefined
\makeatother
\usepackage[hidelinks]{hyperref}

\title{\LARGE \bf
What's Wrong with the Absolute Trajectory Error?
}

\author{Seong Hun Lee and Javier Civera
\thanks{This work was partially supported by the Spanish govt. (PGC2018- 096367-B-I00) and the Arag{\'{o}}n regional govt. (DGA{\_}FSE T45{\_}20R).}
\thanks{Authors are with I3A, University of Zaragoza, 50018 Zaragoza, Spain
{\tt\small seonghunlee@unizar.es; jcivera@unizar.es}}%
}

\begin{document}

\maketitle
\thispagestyle{empty}
\pagestyle{empty}

\begin{abstract}
One of the limitations of the commonly used Absolute Trajectory Error (ATE) is that it is highly sensitive to outliers.
As a result, in the presence of just a few outliers, it often fails to reflect the varying accuracy as the inlier trajectory error or the number of outliers varies.
In this work, we propose an alternative error metric for evaluating the accuracy of the reconstructed camera trajectory.
Our metric, named Discernible Trajectory Error (DTE), is computed in five steps:
(1) Shift the ground-truth and estimated trajectories such that both of their geometric medians are located at the origin.
(2) Rotate the estimated trajectory such that it minimizes the sum of geodesic distances between the corresponding camera orientations.
(3) Scale the estimated trajectory such that the median distance of the cameras to their geometric median is the same as that of the ground truth. 
(4) Compute, winsorize and normalize the distances between the corresponding cameras.
(5) Obtain the DTE by taking the average of the mean and the root-mean-square (RMS) of the resulting distances.
This metric is an attractive alternative to the ATE, in that it is capable of discerning the varying trajectory accuracy as the inlier trajectory error or the number of outliers varies.
Using the similar idea, we also propose a novel rotation error metric, named Discernible Rotation Error (DRE), which has similar advantages to the DTE.
Furthermore, we propose a simple yet effective method for calibrating the camera-to-marker rotation, which is needed for the computation of our metrics. 
Our methods are verified through extensive simulations.
\end{abstract}


\begin{table*}[t]
\begin{center}
\begin{tabular}{c|p{26em}|p{27em}|}
\cline{2-3}
 & \rule{0pt}{1em} \hspace{6em} Interpretation based on the & \hspace{6em}  Interpretation based on the view \\
 & \hspace{6em} transformation of a 3D point & \hspace{6.5em} from another reference frame   \\
\hline
\multicolumn{1}{|c|}{\rule{0pt}{1.2em}$\mathbf{R}_{ij}$} & The rotation that, together with the translation $\mathbf{t}_{ij}$, transforms a 3D point from the reference frame $j$ to $i$: $\mathbf{x}_i=\mathbf{R}_{ij}\mathbf{x}_j+\mathbf{t}_{ij}$.  &  The orientation of the reference frame $j$ with respect to frame $i$.
In other words, the three columns of $\mathbf{R}_{ij}$ correspond to the directions of the standard basis of frame $j$ viewed in frame $i$.\\[3ex]
\hline
\multicolumn{1}{|c|}{\rule{0pt}{1em}$\mathbf{t}_{ij}$} & The translation that, together with the rotation $\mathbf{R}_{ij}$, transforms a 3D point from the reference frame $j$ to $i$: $\mathbf{x}_i=\mathbf{R}_{ij}\mathbf{x}_j+\mathbf{t}_{ij}$. & The position of the reference frame $j$ in frame $i$. 
More specifically, it is the position of the origin of frame $j$ viewed in frame $i$.\\[3ex]
\hline
\end{tabular}
\end{center}
\vspace{-0.5em}
\caption{
Different interpretations of $\mathbf{R}_{ij}$ and $\mathbf{t}_{ij}$.
}
\label{tab:interpretations}
\end{table*}

\section{Introduction}
\label{sec:intro}
Reconstructing a set of camera poses from images (and other sensors) is an important problem in computer vision and robotics.
It has direct application to autonomous navigation, photogrammetry, and AR/VR.
For this reason, significant research endeavors have been devoted to improving the performance of reconstruction algorithms.
Active research areas include odometry \cite{dso, leutenegger_2015_keyframe, loam}, simultaneous localization and mapping (SLAM) \cite{lcsd, ORBSLAM3_TRO, droid}, visual localization \cite{sattler_2018_benchmarking, lynen_2020_ijrr, toft_2022_long} and structure-from-motion  \cite{agarwal_2009_iccv, sfm_revisited, moulon_2013_iccv}.

In pursuit of developing better reconstruction algorithms, it is also very important to ask what error metric should be used to evaluate the accuracy of the results.
If the ground-truth data for the camera poses is given, the current \textit{de facto} standard in the robotics community is the Absolute Trajectory Error (ATE).
An early work that analyzed this metric is \cite{sturm_2012_benchmark}.
Since the ATE is a single number metric that could be intuitively understood and easily used for comparison, it has quickly become a popular choice of metric for the the evaluation of camera localization systems. 

The basic idea of the ATE is to translate, rotate, and optionally, scale the estimated trajectory (\textit{i.e.}, 3D positions of the cameras), such that it is as closely aligned to the ground-truth trajectory as possible.
This is done by minimizing the root mean square (RMS) of the distances between the corresponding cameras in each trajectory, typically using Horn's \cite{horn_1987_closed} and Arun's \cite{arun_1987_least} methods.
The ATE is then given by the optimal RMS value.

Note that the trajectory alignment relies on the optimization under the $L_2$ norm.
This means that the ATE is inevitably sensitive to outliers.
This, in and of itself, is not really a problem, because we want the error metric to clearly indicate whether or not any localization failure has occurred throughout the trajectory.
That said, the actual problem is that, with just a few outliers, it quickly starts losing its sensitivity to the inlier trajectory error and the number of outliers.
In other words, the ATE reacts much less sharply to the changes in these two factors when the estimation contains just a few percent of outliers.

For benchmarking, this is certainly not a desirable property as an error metric.
For example, suppose we are comparing several localization methods on the same dataset and all of the methods happen to produce grossly erroneous estimates at the same set of locations in the trajectory.
In this case, it would be a loss if we cannot know which methods are more accurate within the ``good" part of the trajectory.
Also, if some of the methods fail more times than others, we would want to know in which of them this happens.
Such information is even more relevant when developing one's own system.
Suppose that a baseline algorithm fails to reconstruct some of the camera poses because the dataset involves difficult scenarios.
Suppose we make certain changes in the algorithm that enhance its overall accuracy, but we are still unable to save it from failing. 
In this case, we would want the error metric to reflect the overall increase in accuracy, even if there is no change in  robustness.
In this regard, ATE may not be the most suitable choice of metric.

In this work, we propose a novel alternative to the ATE that addresses this problem. 
Our metric, called Discernible Trajectory Error (DTE), is computed using robust trajectory alignment based on the idea of median.
We show that, unlike the ATE, the DTE can reliably discern the varying accuracy as the inlier trajectory error or the number of outliers varies.
Also, we extend this idea to rotations and propose a novel rotation error metric called Discernible Rotation Error (DRE), which has similar advantages to the DTE.
Furthermore, we propose a simple method for calibrating the camera-to-marker rotation, which needs to be known for the computation of our metrics. 
Our code will be publicly available at {\color{magenta}\url{https://github.com/sunghoon031/DTE_DRE}}.

\section{Related Work}
An early work by Sturm et al. \cite{sturm_2012_benchmark} provides a discussion comparing the ATE and the relative error.
The relative error can be useful for odometry systems \cite{kummerle_2009_auro, burgard_2009_iros, geiger_2012_cvpr}, but it cannot be used for evaluating the reconstruction from unordered image collections.
Zhang and Scaramuzza \cite{zhang_2018_iros} analyze the properties of the ATE and the relative errors for visual(-inertial) systems.
They point out that the ATE decreases when more cameras are used for its computation.
In \cite{zhang_2018_icra}, the same authors propose a continuous-time approach for trajectory evaluation, tackling the potential issue of imperfect temporal association between the ground truth and the estimation.

To our knowledge, in the field of quantitative trajectory evaluation, no previous study has addressed the sensitivity issue caused by outliers (as described in Section \ref{sec:intro}).

\section{Notation}
\label{sec:notation}
For a 3D vector $\mathbf{v}$, we denote its Euclidean norm by $\lVert \mathbf{v} \rVert$ and its unit vector by $\widehat{\mathbf{v}}=\mathbf{v}/\lVert \mathbf{v} \rVert$.
When $\mathbf{v}$ involves physical measurements, we denote it by $\widetilde{\mathbf{
v}}$.
If a rotation matrix $\mathbf{R}$ has the angle $\theta$ and the unit axis of rotation $\widehat{\mathbf{v}}$, then $\mathbf{R}=\mathrm{Exp}(\theta\widehat{\mathbf{v}})$ where $\mathrm{Exp}(\cdot)$ is the mapping defined by Rodrigues' formula \cite{forster_2017_tro}.
The geodesic (or angular) distance between two rotation matrices $\mathbf{R}_j$ and $\mathbf{R}_k$ is defined as the angle of the rotation $\mathbf{R}_j\mathbf{R}_k^\top$ and is denoted by $d(\mathbf{R}_j, \mathbf{R}_k)$.
Let $\mathbf{p}_i$ be a 3D point in the reference frame $i$.
When this point is viewed in another reference frame $j$, its coordinates are given by $\mathbf{p}_j=s_{ji}\mathbf{R}_{ji}\mathbf{p}_i+\mathbf{t}_{ji}$, where $s_{ji}\in\mathbb{R}$, $\mathbf{R}_{ji}\in\mathbf{SO}(3)$ and $\mathbf{t}_{ji}\in\mathbb{R}^3$ are respectively the relative scale, the rotation matrix and the translation vector that relate the reference frame $i$ and $j$.
In the reference frame $j$, the position of the reference frame $i$ is defined as the position of its origin, which is given by $s_{ji}\mathbf{R}_{ji}\mathbf{0}+\mathbf{t}_{ji} = \mathbf{t}_{ji}$.
Likewise, the position of the reference frame $j$ in frame $i$ is given by $\mathbf{t}_{ij}$.
We define the orientation of the reference frame $i$ with respect to frame $j$ as follows:
Imagine three arrows fixed to the reference frame $i$, each pointing in the positive x, y and z direction of frame $i$. 
Let $\mathbf{x}_j$, $\mathbf{y}_j$ and $\mathbf{z}_j$ be the vectors representing these arrows viewed in the reference frame $j$, \textit{i.e.},
$\mathbf{x}_j=s_{ji}\mathbf{R}_{ji}[1,0,0]^\top$, $\mathbf{y}_j=s_{ji}\mathbf{R}_{ji}[0,1,0]^\top$ and 
$\mathbf{z}_j=s_{ji}\mathbf{R}_{ji}[0,0,1]^\top$.
Then, the orientation of the reference frame $i$ with respect to frame $j$ is defined by $\left[\widehat{\mathbf{x}}_j, \widehat{\mathbf{y}}_j, \widehat{\mathbf{z}}_j\right]=\mathbf{R}_{ji}\mathbf{I}_{3\times3}=\mathbf{R}_{ji}$.
Likewise, the orientation of the reference frame $j$ with respect to $i$ is given by $\mathbf{R}_{ij}$.
For clarity, we summarize the notations in Tab. \ref{tab:interpretations}.

\begin{figure}[t]
    \centering
    \includegraphics[width=0.4\textwidth]{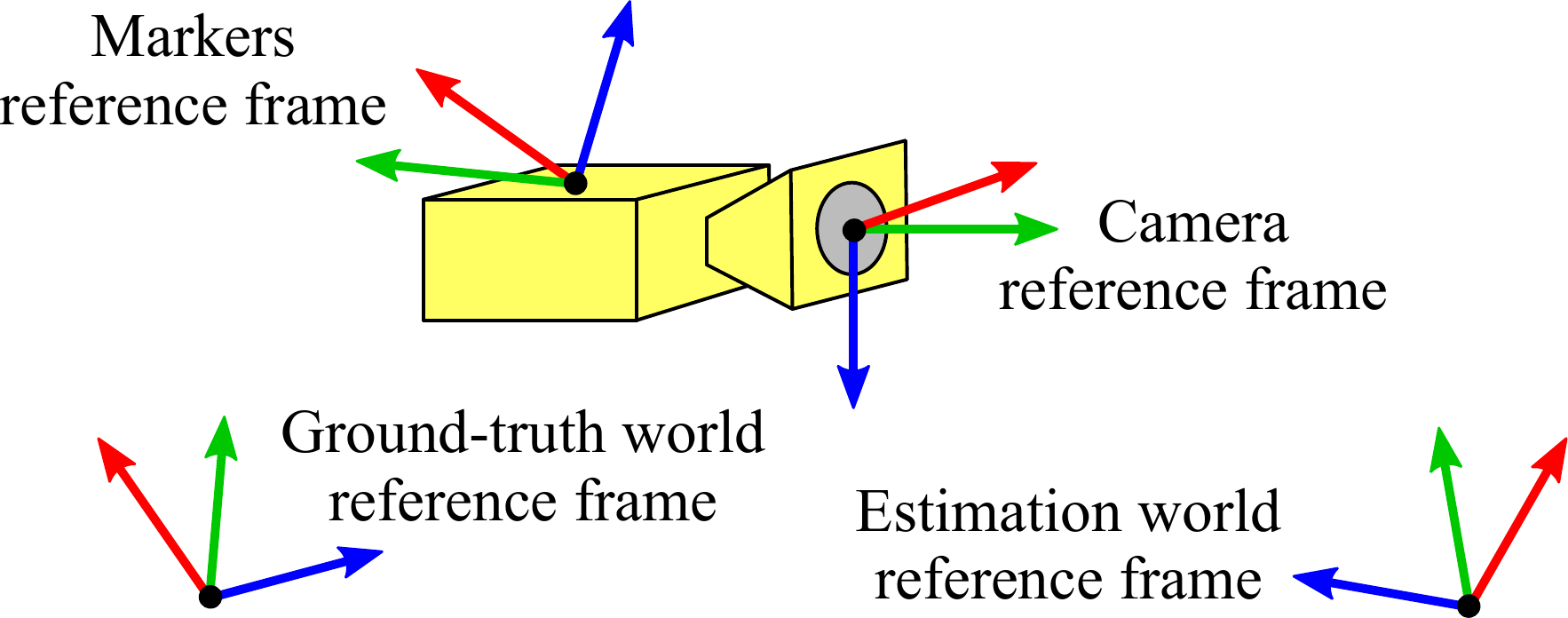}
    \caption{
    Reference frames that are relevant for describing the camera pose.
    }
    \label{fig:frames}
\end{figure}

\section{Preliminaries: ATE}
Suppose that we have measured the 3D positions of $n$ cameras in the ground-truth world reference frame, \textit{i.e.}, $\widetilde{\mathbf{t}}_{gc,i}$ for $i= 1, 2, \cdots, n$.
Also, suppose that we have used some computer vision algorithm and obtained the up-to-scale\footnote{In this work, we assume that the positions are estimated up to scale. 
For applications where the positions are estimated at an absolute scale, one can simply fix the relative scale to 1 in the remainder of the paper.} positions of those cameras in the estimation world reference frame, \textit{i.e.}, $\mathbf{t}_{ec,i}$ for $i= 1, 2, \cdots, n$.
Now, the ATE between those two sets of camera positions is defined as follows \cite{sturm_2012_benchmark}:
\begin{equation}
\label{eq:ATE}
    \text{ATE}=\min_{\scaleto{s_{ge}, \mathbf{R}_{ge}, \mathbf{t}_{ge}}{8pt}}\sqrt{\frac{1}{n}\sum_{i=1}^n\lVert\widetilde{\mathbf{t}}_{gc, i}-\left(s_{ge}\mathbf{R}_{ge}\mathbf{t}_{ec, i} + \mathbf{t}_{ge}\right)\rVert^2}.
\end{equation}
Essentially, this is the minimum RMS of the distances between the ground-truth trajectory and the estimated trajectory after aligning the latter to the former using $\mathbf{SIM}(3)$ transformation. 

Since it is impossible to measure $\widetilde{\mathbf{t}}_{gc}$ directly, we use reflective markers (or other measuring tools) instead to track the position of the cameras.
As a result, the reference frame defined by the markers may not be exactly the same as the camera reference frame (see Fig. \ref{fig:frames}).
These two reference frames are related by the following transformation of a 3D point:
$\mathbf{p}_m = s_{mc}\mathbf{R}_{mc}\mathbf{p}_c + \mathbf{t}_{mc}$.
Combining this with the transformation between the markers and the ground-truth world reference frame, we get
\begin{align}
    \mathbf{p}_g&=\mathbf{R}_{gm}\mathbf{p}_m+\mathbf{t}_{gm} \\ 
    &=s_{mc}\underbrace{\mathbf{R}_{gm}\mathbf{R}_{mc}}_{\scaleto{\mathbf{R}_{gc}}{8pt}}\mathbf{p}_c+\underbrace{\mathbf{R}_{gm}\mathbf{t}_{mc}+\mathbf{t}_{gm}}_{\scaleto{\mathbf{t}_{gc}}{8pt}}.
\end{align}
This means that $\widetilde{\mathbf{t}}_{gc, i}$ in \eqref{eq:ATE} can be obtained as follows:
\begin{equation}
\label{eq:t_gc}
    \widetilde{\mathbf{t}}_{gc, i} = \widetilde{\mathbf{R}}_{gm,i}\mathbf{t}_{mc}+\widetilde{\mathbf{t}}_{gm,i}
\end{equation}
In practice, if $\mathbf{t}_{mc}$ is unknown, we often assume that it is zero, as the markers are usually placed near the camera\footnote{It would be problematic to assume that $\mathbf{t}_{mc}$ is zero when it is non-negligible.
Here is a simple example:
Imagine we have a long rod.
We attach the camera at one end, A, and the markers at the other end, B.
Now, suppose that we move A in a circle while keeping B at the center of the circle.
Replacing $\widetilde{\mathbf{t}}_{gc, i}$ with $\widetilde{\mathbf{t}}_{gm, i}$ in \eqref{eq:ATE}, we get $\text{ATE}=0$ because $\widetilde{\mathbf{t}}_{gm, i}=\mathbf{0}$, and thus, $s_{ge}=0$ and $\mathbf{t}_{ge}=\mathbf{0}$. 
This does not happen if we use \eqref{eq:t_gc} without assuming that $\mathbf{t}_{mc}=\mathbf{0}$.}.

In \cite{horn_1987_closed} and \cite{arun_1987_least}, the closed-form optimal solution for \eqref{eq:ATE} was derived.
We refer to the original works for the derivations.
In the following, we summarize the steps needed to obtain the solution:
\begin{enumerate}
    \item Compute the centroids of the two trajectories (\textit{i.e.}, the ground truth, $\widetilde{\mathbf{t}}_{gc, i}$, and the estimation, $\mathbf{t}_{ec, i}$ for all $i$):
    \begin{equation}
        \overline{\mathbf{t}}_{gc} = \frac{\sum_{i=1}^n \widetilde{\mathbf{t}}_{gc, i}}{n}, \quad 
        \overline{\mathbf{t}}_{ec} = \frac{\sum_{i=1}^n \mathbf{t}_{ec, i}}{n},
    \end{equation}
    Let $\mathbf{G}$ and $\mathbf{E}$ be the $3\times n$ matrices that have 
    $\widetilde{\mathbf{t}}_{gc, i}-\overline{\mathbf{t}}_{gc}$ and $\mathbf{t}_{ec, i}-\overline{\mathbf{t}}_{ec}$ as their columns, respectively.
    \item Compute the singular value decomposition of $\mathbf{EG}^\top$:
     \begin{equation}
        \mathbf{U\Sigma V}^\top = \mathrm{SVD}\left(\mathbf{EG}^\top\right).
    \end{equation}
    Then, the optimal rotation $\mathbf{R}_{ge}$ is given by 
    \begin{equation}
    \label{eq:ATE_rotation}
        \mathbf{R}_{ge}^*=\mathbf{V}
        \begin{bmatrix}
        1 & 0 & 0 
        \\ 0 & 1 & 0 
        \\ 0 & 0 & \text{sign}\left(\text{det}\left(\mathbf{VU^\top}\right)\right) 
        \end{bmatrix}
        \mathbf{U}^\top.
    \end{equation}
    \item The optimal relative scale $s_{ge}$ is given by
    \begin{equation}
    \label{eq:ATE_scale}
        s_{ge}^*= \frac{\sum_{i=1}^n\mathbf{G}_i^\top\mathbf{R}_{ge}^*\mathbf{E}_i}{\sum_{i=1}^n \mathbf{E}_i^\top\mathbf{E}_i},
    \end{equation}
    where $\mathbf{G}_i$ and $\mathbf{E}_i$ denote the $i$th column of each matrix.
    \item The optimal translation $\mathbf{t}_{ge}$ is given by
    \begin{equation}
    \label{eq:ATE_translation}
        \mathbf{t}_{ge}^* = \overline{\mathbf{t}}_{gc}-s_{ge}^*\mathbf{R}_{ge}^*\overline{\mathbf{t}}_{ec}.
    \end{equation}
\end{enumerate}
The ATE is then obtained by plugging these results into \eqref{eq:ATE}.
Note that each of these steps is vulnerable to potential outliers in the estimated trajectory.
In the next section, we propose an alternative metric that can better handle outliers.

\section{Discernible Trajectory Error (DTE)}
\label{sec:DTE}
The key idea behind our method is to robustify each step in aligning the two trajectories.
The proposed steps are as follows:
\begin{enumerate}
    \item Compute the geometric medians of the two trajectories (\textit{i.e.}, the ground truth, $\widetilde{\mathbf{t}}_{gc, i}$, and the estimation, $\mathbf{t}_{ec, i}$ for all $i$):
    \begin{align}
        \mathbf{t}_{gc}^* &= \argmin_{\scaleto{\mathbf{t}}{5pt}}\sum_{i=1}^n \ \lVert\widetilde{\mathbf{t}}_{gc,i}-\mathbf{t}\rVert, \\
        \mathbf{t}_{ec}^* &= \argmin_{\scaleto{\mathbf{t}}{5pt}}\sum_{i=1}^n \ \lVert\mathbf{t}_{ec,i}-\mathbf{t}\rVert.
    \end{align}
    These can be found efficiently using the Weiszfeld algorithm \cite{weiszfeld1, weiszfeld2}.
    \item Find the optimal rotation $\mathbf{R}_{ge}$ that rotates the estimated trajectory such that it minimizes the sum of geodesic distances between the corresponding camera orientations in each trajectory:
    \begin{equation}
    \label{eq:DTE_rotation1}
        \mathbf{R}_{ge}^* =  \argmin_{\mathbf{R}_{\scaleto{\textrm{{align}}}{6pt}}}\sum_{i=1}^n d\left(\widetilde{\mathbf{R}}_{gc,i}, \mathbf{R}_{\text{align}}\mathbf{R}_{ec,i}\right).
    \end{equation}
    Since the geodesic distance is invariant to rotation, we can rearrange this equation into the following form:
    \begin{equation}
    \label{eq:DTE_rotation2}
        \mathbf{R}_{ge}^* =  \argmin_{\mathbf{R}_{\scaleto{\textrm{{align}}}{6pt}}}\sum_{i=1}^n d\left(\widetilde{\mathbf{R}}_{gc,i}\mathbf{R}_{ec,i}^\top, \mathbf{R}_{\text{align}}\right).
    \end{equation}
    This is the single rotation averaging problem on $\mathbf{SO}(3)$ under the $L_1$ norm, and $\mathbf{R}_{ge}^*$ corresponds to the geodesic median of the rotations $\widetilde{\mathbf{R}}_{gc,i}\mathbf{R}_{ec,i}^\top$ for all $i$.
    We solve this using the method proposed in \cite{hartley_2011_l1}.
    Note that the ground-truth camera orientations $\widetilde{\mathbf{R}}_{gc,i}$ are obtained as follows:
    \begin{equation}
    \label{eq:DTE_Rmc}
        \widetilde{\mathbf{R}}_{gc,i}=\widetilde{\mathbf{R}}_{gm,i}\widetilde{\mathbf{R}}_{mc},
    \end{equation}
    where $\widetilde{\mathbf{R}}_{mc}$ is the rotation between the camera and the markers reference frame.
    For now, we assume that this is already known (see Section \ref{sec:calibration} for its calibration).
    
    \item Compute the relative scale $s_{ge}$ as follows:
    \begin{equation}
    \label{eq:DTE_scale}
        s_{ge} = \frac{\underset{i}{\mathrm{med}} \ \lVert\widetilde{\mathbf{t}}_{gc,i}-\mathbf{t}_{gc}^*\rVert}{{\underset{i}{\mathrm{med}} \ \lVert\mathbf{t}_{ec, i}}-\mathbf{t}_{ec}^* \rVert}.
    \end{equation}
    Note that the numerator and the denominator correspond to the median absolute deviations (MAD) \cite{hampel_1974_influence} to the geometric median of the ground-truth and estimated camera positions, respectively.
    
    \item Compute the translation $\mathbf{t}_{ge}$ as follows:
    \begin{equation}
    \label{eq:DTE_translation}
        \mathbf{t}_{ge} = \mathbf{t}_{gc}^* - s_{ge}\mathbf{R}_{ge}^*\mathbf{t}_{ec}^*.
    \end{equation}
    
    \item Compute the distances between the corresponding cameras and winsorize them such that none of the distances is larger than $k$ times the MAD to the ground-truth median.
    Then, divide the results by this upper bound, so that the theoretical maximum becomes 1:
    For $i = 1, \cdots, n$,
    \begin{align}
        d_i &= \lVert\widetilde{\mathbf{t}}_{gc, i}-\left(s_{ge}\mathbf{R}_{ge}^*\mathbf{t}_{ec, i} + \mathbf{t}_{ge}\right)\rVert, \\
        \epsilon_i &= \frac{\mathrm{min}\left(d_i, \ u \right)}{u}
        \quad \text{with} \quad  u=k\times\underset{i}{\mathrm{med}} \ \lVert \widetilde{\mathbf{t}}_{gc,i}-\mathbf{t}_{gc}^*\rVert \label{eq:winsorized_error}
    \end{align}
    The larger the parameter $k$ in \eqref{eq:winsorized_error}, the greater the influence of outliers. 
    For the experiments in this work, we found that setting $k=5$ was appropriate.
    
    \item The DTE is then defined by
    \begin{equation}
    \label{eq:DTE_final}
        \text{DTE} = \frac{1}{2}\left(\sum_{i=1}^n \ \frac{\epsilon_i}{n} + \sqrt{\sum_{i=1}^n \ \frac{\left(\epsilon_i\right)^2}{n}} \ \   \right).
    \end{equation}
    
    That is, the DTE is the average of the mean and the RMS of the trajectory errors after aligning the estimated trajectory using the transformation obtained from the previous steps.
    Note that the DTE is a dimensionless metric that always lies between 0 and 1 as a result of the winsorization and normalization performed in \eqref{eq:winsorized_error}.
\end{enumerate}

In the following, we provide the rationale behind our approach:
First, we use the geometric median of the trajectory instead of the centroid when computing the translation.
This is because the geometric median is much more robust to outliers than the centroid \cite{lopuhaa_1991_breakdown}.
As a result, our translation \eqref{eq:DTE_translation} is more robust than that used for the ATE \eqref{eq:ATE_translation}.

Second, we obtain the rotation by aligning the camera orientations using the geodesic median in $\mathbf{SO}(3)$.
For the same reason as previously stated, this is more robust than \eqref{eq:ATE_rotation} which assigns equal weights to both inliers and outliers.

Third, we compute the scale using the MAD, instead of the variance (as in \eqref{eq:ATE_scale}) or the standard deviation (as in the symmetrical scale proposed in \cite{horn_1987_closed}).
The inherent robustness of the MAD \cite{hampel_1974_influence} makes our scale estimate \eqref{eq:DTE_scale} more reliable than \eqref{eq:ATE_scale} which is highly sensitive to outliers.

Fourth, we winsorize the trajectory errors to prevent the unbounded error.
While a single extreme outlier would render the ATE useless, the DTE can handle any number of unbounded outliers owing to this measure.

Fifth, we normalize the winsorized errors by dividing them by their upper bound $u$ in \eqref{eq:winsorized_error}.
This prevents situations where the winsorization hides large errors in a small-scale dataset. 
For instance, if the ground-truth trajectory is very small in magnitude, $u$ in \eqref{eq:winsorized_error} would be very small too, no matter how inaccurate the estimation is.
In this case, without the normalization, the resulting DTE would be very close to $u$.
This could be misinterpreted as an accurate estimation result.
The division by $u$ resolves this issue, as the DTE would be close to 1, the maximum possible value.

Finally, instead of simply taking the mean or the RMS error after alignment, we take their average in \eqref{eq:DTE_final}.
This way, the DTE behaves favorably in terms of sensitivity to the inlier trajectory accuracy and the number of outliers.
We elaborate on this behavior in Section \ref{subsec:result_why}.

\section{Discernible Rotation Error (DRE)}
We can apply the similar principle when evaluating the accuracy of the rotations separately:
We define the Discernible Rotation Error (DRE) as the average of the mean and RMS rotation error after aligning the camera orientations using the same rotation $\mathbf{R}_{ge}^*$ defined in \eqref{eq:DTE_rotation1}.
Like the DTE, the DRE is capable of discerning the varying accuracy as the inlier orientation error or the number of outliers varies.
Since rotation errors are bounded unlike translation errors, we do not perform winsorization or normalization.
The evaluation results are discussed in Section \ref{subsec:result_DRE}.

\section{Calibration of Camera-to-Marker Rotation}
\label{sec:calibration}
So far, we have assumed that the camera-to-marker rotation $\widetilde{\mathbf{R}}_{mc}$ in \eqref{eq:DTE_Rmc} is known.
In this section, we propose a simple method for calibrating this rotation.
The idea is to optimize two rotations simultaneously by solving the following problem:
\begin{equation}
\label{eq:calibration}
    \argmin_{\mathbf{R}_{\scaleto{\textrm{{align}}}{6pt}}, \  \widetilde{\mathbf{R}}_{\scaleto{mc}{3pt}}} \ 
    \sum_{i=1}^n d\left(\widetilde{\mathbf{R}}_{gm,i}\widetilde{\mathbf{R}}_{mc}\mathbf{R}_{ec,i}^\top, \mathbf{R}_{\text{align}}\right).
\end{equation}
Notice that this is a simple extension of \eqref{eq:DTE_rotation2} which now involves another unknown variable, $\widetilde{\mathbf{R}}_{mc}$.
Assuming that the estimated rotations $\mathbf{R}_{ec,i}$ are reasonably accurate and that the camera orientations are not in any degenerate configurations (which will be discussed later), solving this problem will lead to an accurate estimate of $\widetilde{\mathbf{R}}_{mc}$ (and $\mathbf{R}_{\text{align}}$ if needed).

Our strategy for solving \eqref{eq:calibration} is as follows:

\begin{enumerate}\itemsep5pt
    \item Set $\mathbf{R}_\mathrm{est}=\mathbf{I}_{3\times3}$ and $\theta_\mathrm{max}=360^\circ$.
    
    \item Set $\theta = k\theta_\mathrm{max}$ where $k$ is a random number between 0 and 1.
    Also, set $\widehat{\mathbf{v}}$ to a random unit vector.
    
    \item Update $\widetilde{\mathbf{R}}_{mc}\gets\mathrm{Exp}(\theta\widehat{\mathbf{v}})\mathbf{R}_\mathrm{est}$.
    
    \item Since $\widetilde{\mathbf{R}}_{mc}$ is now fixed, \eqref{eq:calibration} becomes the single rotation averaging problem on $\mathbf{SO}(3)$ under the $L_1$ norm.
    Solve it using the method proposed in \cite{hartley_2011_l1}.
    
    \item Repeat Step 2--4 one thousand times. 
    Whenever \eqref{eq:calibration} yields a smaller cost than the smallest value ever obtained so far, update $\mathbf{R}_\mathrm{est}\gets\widetilde{\mathbf{R}}_{mc}$.
    
    \item Repeat Step 2--5 four more times.
    For each further iteration, update $\theta_\mathrm{max}$ to $30^\circ$, $10^\circ$, $3^\circ$ and $1^\circ$, respectively.
    Return the final $\widetilde{\mathbf{R}}_{mc}$ in the end.
\end{enumerate}
Essentially, we update the estimate of $\widetilde{\mathbf{R}}_{mc}$ by searching for a better rotation within the ball of a certain radius around the current estimate.
In the outer loop, the radius of the ball is decreased at each iteration, and in the inner loop, the search is done by simple random sampling within the ball.

\textbf{A word of caution:}
The calibration method described above assumes that the estimated rotations 
$\mathbf{R}_{ec,i}$ are accurate.
Indeed, our experiment shows that their overall accuracy directly affects the calibration result (see Section \ref{subsec:result_calib}).
That said, since our method is based on the robust rotation averaging algorithm \cite{hartley_2011_l1}, the calibration error will be well within a tolerable limit as long as the estimation is reasonably accurate.
For example, the calibration error is mostly less than $1^\circ$ when the average estimation error is less than $10^\circ$.
To prevent a potential calibration failure in the first place, one may consider performing the calibration procedure in a separate session using, for example, a checkerboard pattern.

Another important aspect that must be taken into account during calibration is that certain camera orientations can lead to degeneracy.
Specifically, this happens when all cameras have the same fixed orientation, or when they all rotate around same axis.
The proof is given in the appendix.

\begin{figure}[t]
    \centering
    \includegraphics[width=0.485\textwidth]{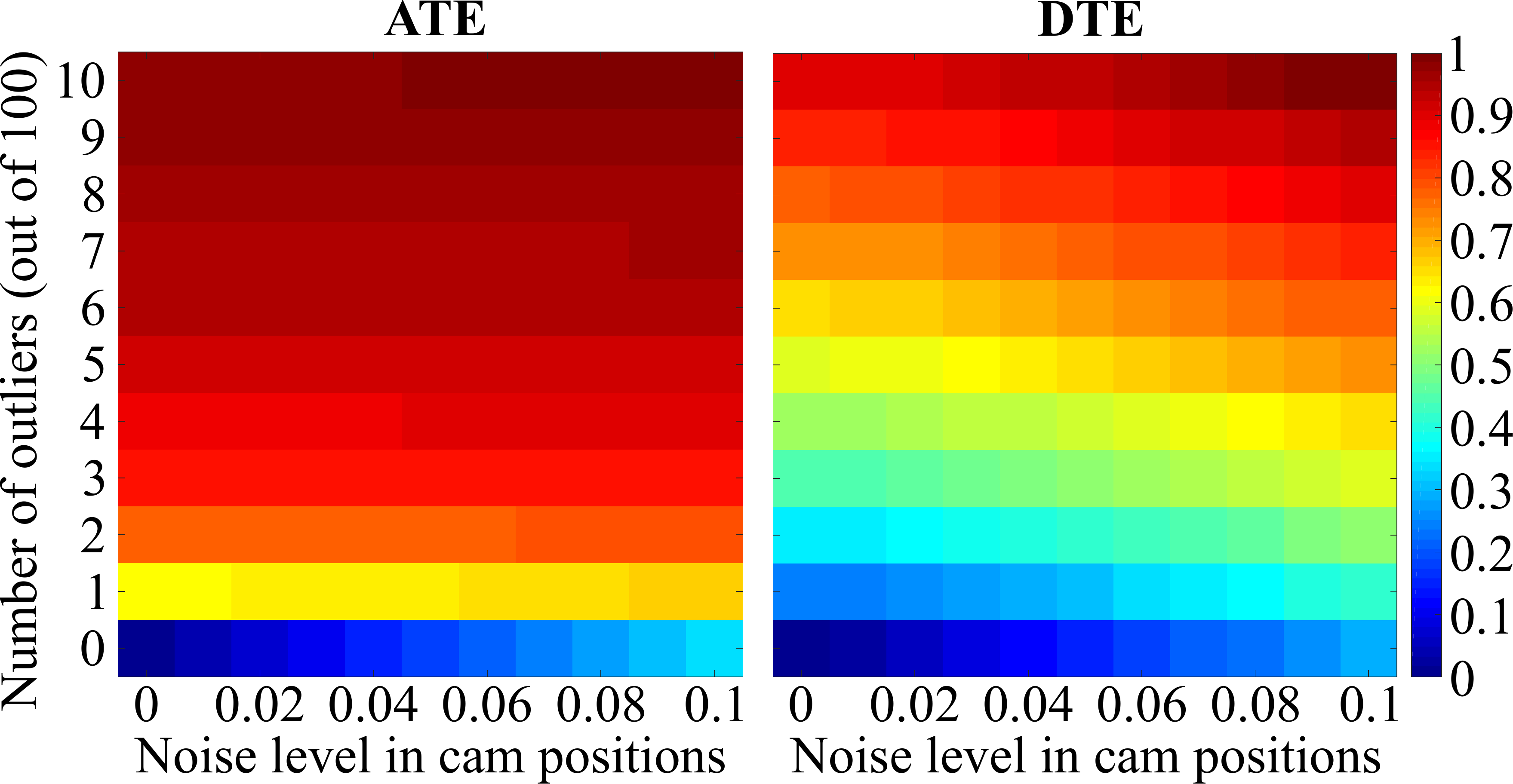}
    \caption{
    Each of the colored blocks represents the normalized ATE and DTE obtained with the given number of outliers and noise level in camera positions.
    In the presence of outliers, the DTE shows a more pronounced gradation than the ATE.
    This means that it is better at capturing the varying trajectory accuracy as the number of outliers and the noise level varies.
    }
    \label{fig:ATE_vs_DTE}
\end{figure}
\begin{figure}[t]
    \centering
    \includegraphics[width=0.485\textwidth]{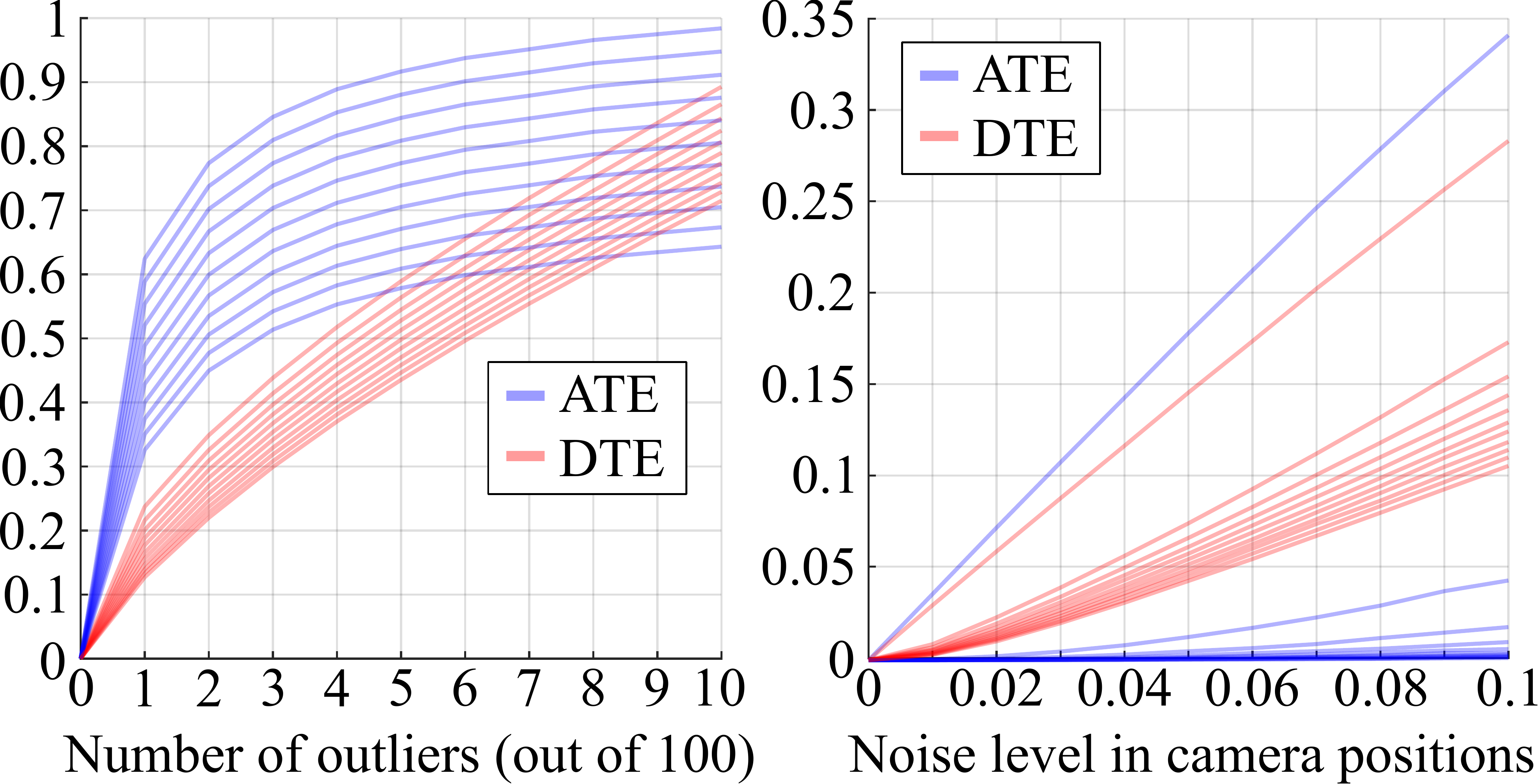}
    \caption{
    \textbf{[Left]} The errors corresponding to each column of Fig. \ref{fig:ATE_vs_DTE}, showing the effect of a varying number of outliers.
    We shift each column values such that the error in the outlier-free case is 0.
    As the number of outliers increases, the ATE curves flatten, which signals a decreasing level of sensitivity.
    In contrast, the DTE maintains a relatively high level of sensitivity.
    \textbf{[Right]} The errors corresponding to each row of Fig. \ref{fig:ATE_vs_DTE}, showing the effect of a varying noise level in camera positions. We shift each row values such that the error in the zero-noise case is 0.
    As the number of outliers increases, the slope of the ATE curve drastically decreases, and with just three outliers, it becomes almost insensitive to the noise level.
    In contrast, the DTE can still maintain a moderate level of sensitivity, even with 10 outliers.
    }
    \label{fig:sensitivity}
\end{figure}

\section{Results}
\subsection{Comparison between ATE and DTE}
\label{subsec:ate_vs_dte}
We compare the ATE and the DTE in simulation:
We generate $100$ cameras with random rotations and positions inside a $1\times1\times1$ cube centered at the origin.
Then, we corrupt this ground truth to obtain the estimated trajectory.
First, we perturb the camera positions with Gaussian noise $\mathcal{N}(0, \sigma^2)$ where $\sigma=0, 0.01, 0.02, \dots, 0.1$ unit.
Their rotations are perturbed by Gaussian noise with $\sigma = 5^\circ$.
Next, we turn some of the cameras into outliers by assigning to them random rotations and positions inside a $10\times10\times10$ cube centered at the origin. 
We vary the number of outliers between 0 and 10 in our experiment.

Next, we rotate the entire trajectory by a random rotation, scale it by a random number, and shift it by a random translation vector.
The resulting trajectory is taken as our estimation.
Note that here we assume that $\widetilde{\mathbf{R}}_{mc}$ in \eqref{eq:DTE_Rmc} is the identity matrix and that this is known in advance through calibration.
We compute the DTE and the ATE by comparing the estimated trajectory with the ground truth.
This procedure gives us errors in 121 settings because we have 11 different numbers of outliers and 11 different noise levels in camera positions.
We repeat this 1000 times with independently generated ground-truth trajectories.

In order to aggregate the results from these 1000 runs, we normalize the 121 ATEs and 121 DTEs from each run by dividing them by the maximum ATE and DTE, respectively.
This gives us, per run, 121 errors up to the magnitude of 1, which can be averaged over the 1000 runs.

\begin{figure}[t]
    \centering
    \includegraphics[width=0.485\textwidth]{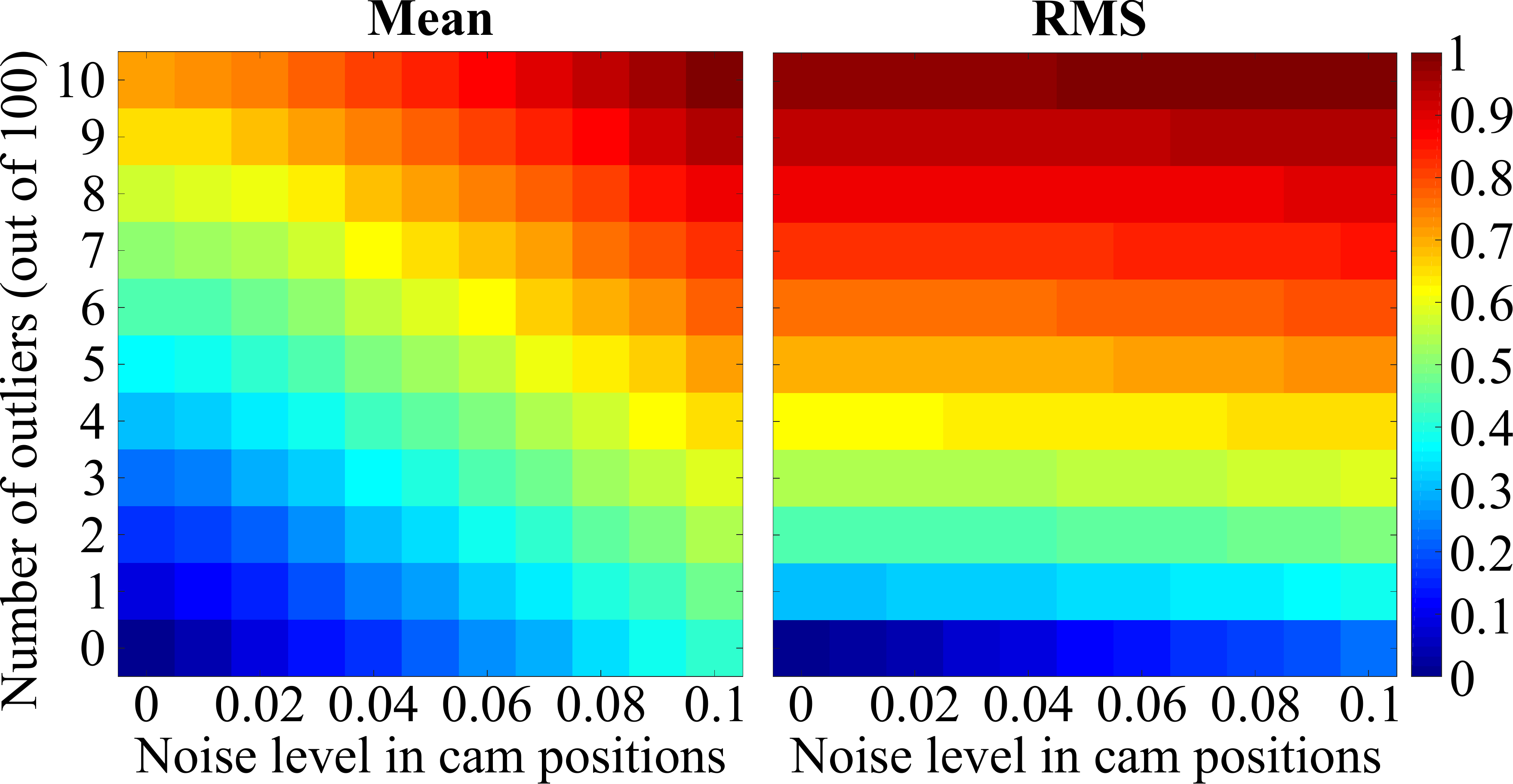}
    \caption{
    Each of the colored blocks represents the normalized mean and RMS trajectory errors obtained with the given number of outliers and noise level in camera positions.
    The mean error shows a high level of sensitivity to both the number of outliers and the noise level, whereas the RMS error is sensitive to the number of outliers, but not so much to the noise level.
    }
    \label{fig:mean_vs_rms}
\end{figure}

Fig. \ref{fig:ATE_vs_DTE} shows the results.
We observe that the ATE quickly becomes insensitive to the number of outliers and the noise level as the outlier ratio increases.
On the other hand, the DTE can discern the effect of a varying number of outliers and noise level relatively well across the whole domain of settings.
This contrast in their sensitivity is more clearly shown in Fig. \ref{fig:sensitivity}, where we plot the error values of each row and column of Fig. \ref{fig:ATE_vs_DTE} separately.

\subsection{Why take the average of the mean and the RMS?}
\label{subsec:result_why}
In this section, we explain why we take the average of the mean and RMS error in \eqref{eq:DTE_final}.
Fig. \ref{fig:mean_vs_rms} shows the results if we choose either only the mean or only the RMS.
In the left panel, we observe that the mean error is sensitive both to the number of outliers and the noise level.
This is not a problem per se, but it is when the error reacts similarly to the change in the number of outliers and the change in the noise level.
For many robotics applications, the estimator's robustness is generally considered more critical than its overall accuracy.
Therefore, we want the error metric to be substantially more sensitive to a camera tracking failure than to a slight loss in accuracy.
For the RMS error, on the other hand, this is not an issue (see the right panel of Fig. \ref{fig:mean_vs_rms}), but it is rather insensitive to the inlier noise level.
To overcome the respective weaknesses of the mean and the RMS error, we combine the two together.
This way, the mean error provides the sensitivity to the noise level, while the RMS error accentuates the effect of outliers.

We point out that \eqref{eq:DTE_final} can be seen as an instance of a more general form involving variable weights, \textit{i.e.,}
\begin{equation}
    (1-\alpha)\epsilon_\mathrm{mean}+\alpha\epsilon_\mathrm{rms},
\end{equation}
with $0\leq\alpha\leq 1$.
The greater the value of $\alpha$, the stronger the influence of outliers.
While the appropriate value for $\alpha$ was $0.5$ in our experiment, it is also possible to choose a different value for other applications, depending on the noise/outlier sensitivity.

\begin{figure*}[t]
    \centering
    \includegraphics[width=0.95\textwidth]{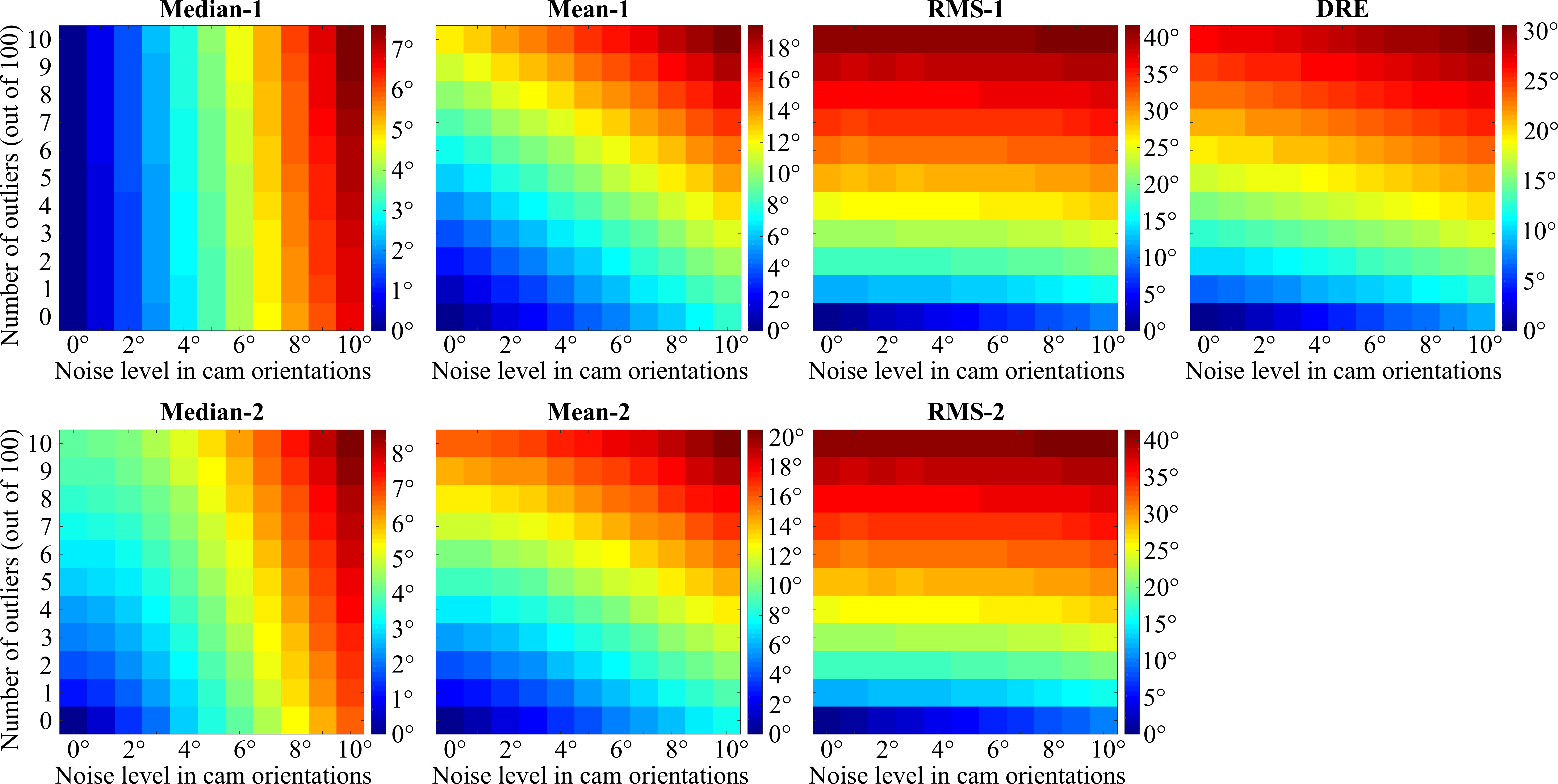}
    \caption{
    Each of the colored blocks represents the rotation error obtained with the given number of outliers and noise level in camera orientations.
    Compared to the RMS errors, the median and mean errors have relatively low sensitivity to the number of outliers and relatively high sensitivity to the noise level.
    The DRE is sufficiently sensitive to the noise level, but more to the number of outliers, which is a desirable property.
    }
    \label{fig:rotation}
\end{figure*}

\subsection{Evaluation of the DRE}
\label{subsec:result_DRE}
We compare the DRE and the following error metrics:
\begin{itemize}
    \item Median-1 \cite{chatterjee_2018_tpami}, Mean-1 \cite{lee_2021_cvpr}, RMS-1: 
    These are respectively the median, mean and RMS rotation error after aligning the orientations by minimizing the geodesic distances under the $L_1$ norm.
    \item Median-2 \cite{chatterjee_2018_tpami}, Mean-2 \cite{lee_2021_cvpr}, RMS-2 \cite{lee_2022_cvpr}:
    These are the counterparts of the previous three, obtained using the alignment under the $L_2$ norm.
\end{itemize}
Note that the DRE is basically the average of Mean-1 and RMS-1 errors.
To compare these error metrics in a controlled manner, we generate 100 random ground-truth orientations and obtain the estimated orientations by (1) rotating the ground truth by some random rotation, (2) perturbing them with Gaussian noise, and (3) turning some of them into outliers by setting them to random orientations.
We vary the noise level up to $10^\circ$ and the outlier ratio up to $10\%$.
For each configuration, we run 1000 independent simulations.
The average results are shown in Fig. \ref{fig:rotation}.
From this figure, we can see that the DRE has similar advantages as the DTE.

\begin{figure}[t]
    \centering
    \includegraphics[width=0.3\textwidth]{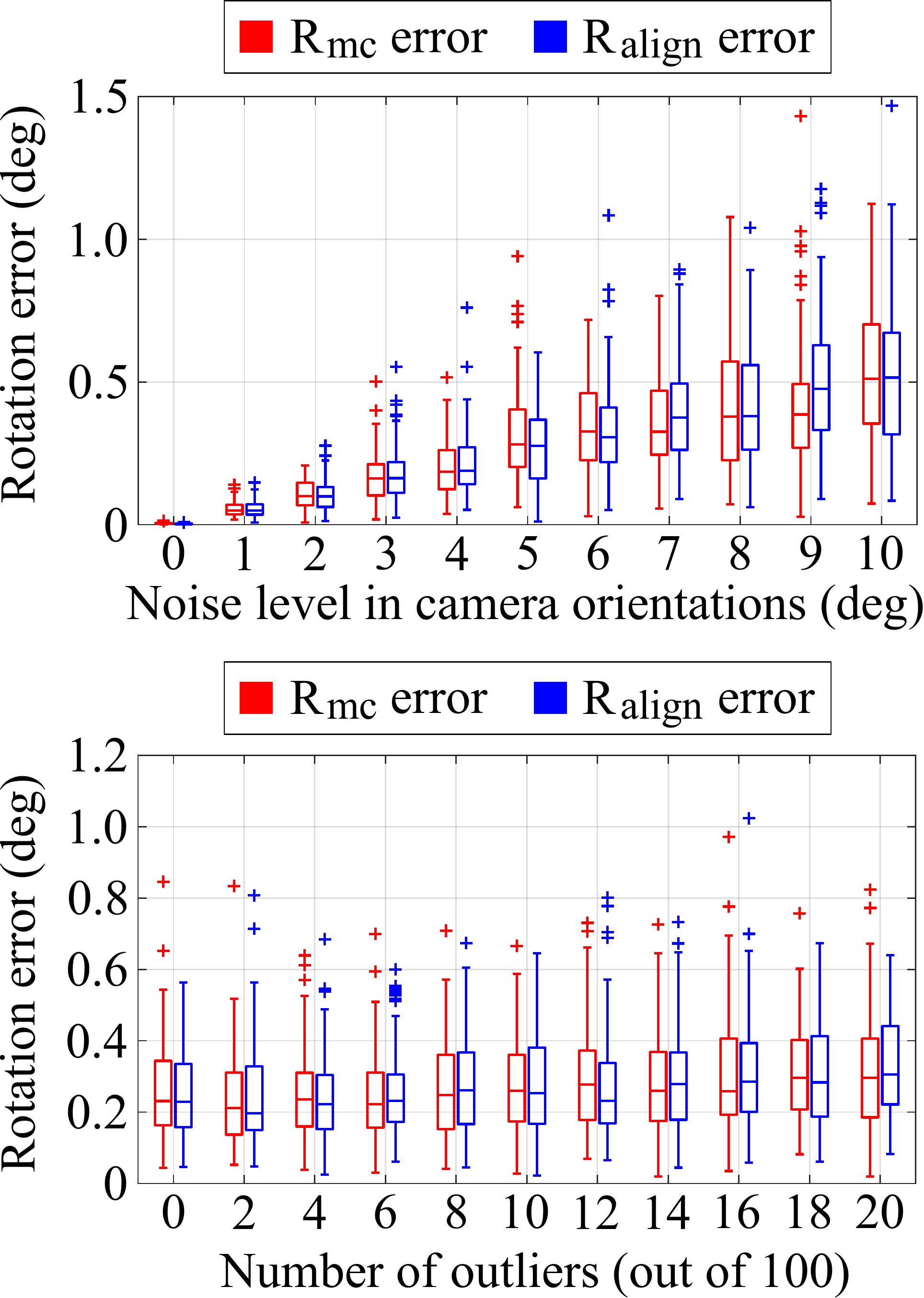}
    \caption{\textbf{[Top]} Calibration errors under a varying noise level in the estimated camera orientations.
    We fix the number of outliers to be 5 out of 100.
    The median error increases somewhat linearly to the noise level.
    \textbf{[Bottom]} Calibration errors under a varying number of outliers in the estimated camera orientations.
    We fix the noise level to be $5^\circ$.
    The effect of outliers is relatively small when the outlier ratio is moderate (\textit{i.e.}, at least up to 20\%).}
    \label{fig:calib_errors}
\end{figure}

\subsection{Evaluation on a real dataset}
We repeat the experiments described in Section \ref{subsec:ate_vs_dte} and \ref{subsec:result_DRE} using the groundtruth data of the \texttt{freiburg2\_desk} sequence from the TUM RGB-D dataset \cite{sturm_2012_benchmark}.
We downsample the sequence with a sampling interval of $0.5s$, resulting in $161$ camera poses.
We then scale and shift the groundtruth trajectory such that it fits inside a $1\times1\times1$ cube centered at the origin, and obtain the estimated trajectory by simulating the noise, outliers and transformation as described in Section \ref{subsec:ate_vs_dte}.
We found that the results are almost indistinguishable from that shown in Fig. \ref{fig:ATE_vs_DTE} and \ref{fig:rotation}, so these redundant figures are omitted in this paper.

\subsection{Evaluation of our calibration method}
\label{subsec:result_calib}
To evaluate our calibration method in Section \ref{sec:calibration}, we set up the simulation as follows:
First, we generate 100 random ground-truth orientations, $\widetilde{\mathbf{R}}_{gm,i}$, and two random rotations, $\mathbf{R}_\text{align}$ and $\widetilde{\mathbf{R}}_{mc}$.
Next, we generate the estimated orientations, $\mathbf{R}_{ec,i}$, by setting them to $\mathbf{R}_{\text{align}}^\top\widetilde{\mathbf{R}}_{gm,i}\widetilde{\mathbf{R}}_{mc}$.
Finally, we perturb these orientations with Gaussian noise and turn some of them into outliers by setting them to random orientations.
We vary the noise level up to $10^\circ$ and the outlier ratio up to $20\%$.
For each setting, we generate 100 independent datasets.
Then, using the ground-truth and estimated orientations as input, we solve \eqref{eq:calibration} and obtain the estimates of $\mathbf{R}_\text{align}$ and $\widetilde{\mathbf{R}}_{mc}$.

\begin{figure}[t]
    \centering
    \includegraphics[width=0.485\textwidth]{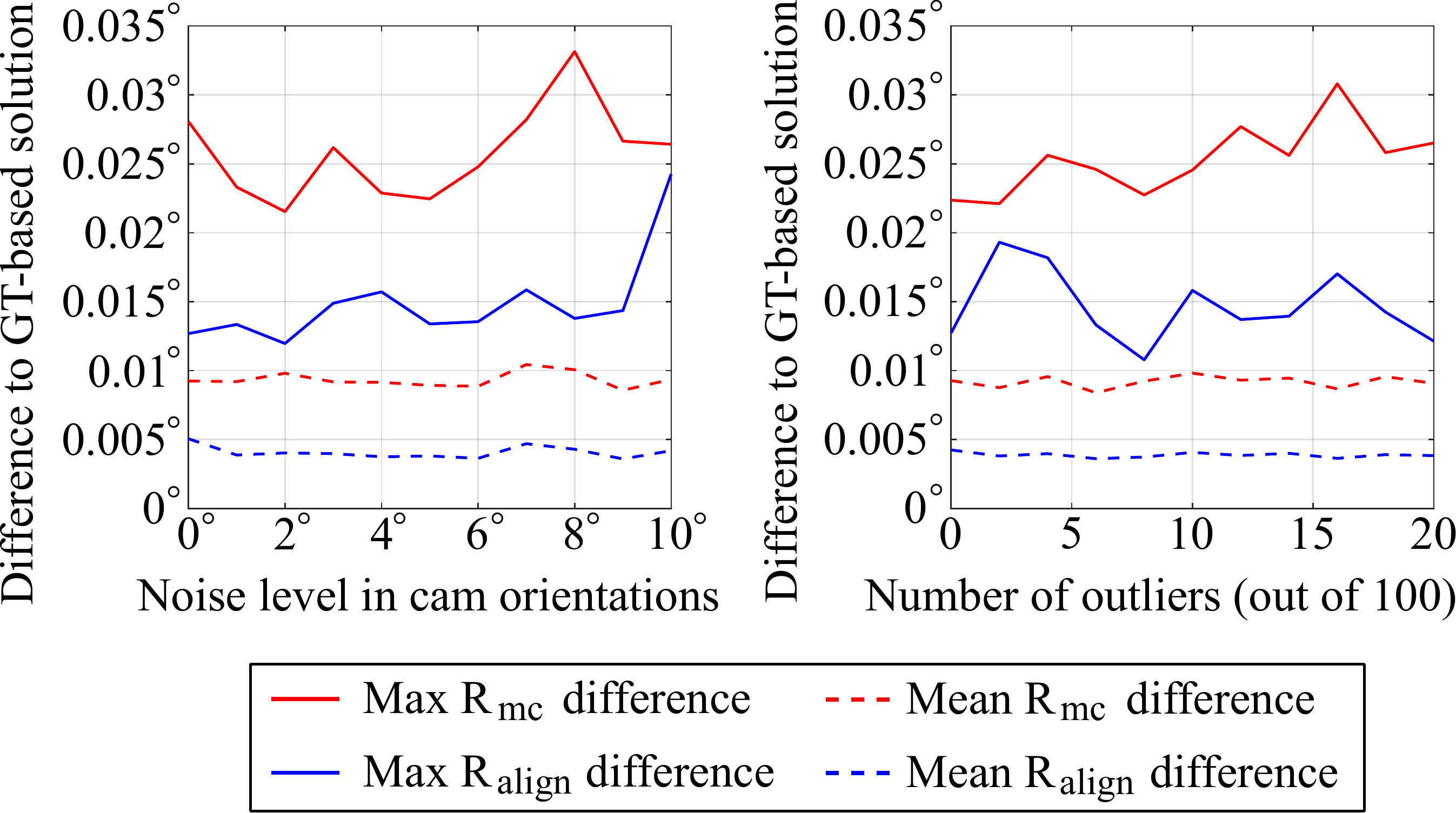}
    \caption{
    The angular difference between our calibration results and the results we get when we use the ground truth as the initial seed:
    \textbf{[Left]} We vary the noise level in the camera orientations, while fixing the number of outliers to be 10 out of 100.
    \textbf{[Right]} We vary the number of outliers, while fixing the noise level in the camera orientations to be 5 deg.
    }
    \label{fig:calib_gt_init}
\end{figure}

Fig. \ref{fig:calib_errors} presents the results.
It shows that the median calibration error is less than $0.5^\circ$ when the estimation is reasonably accurate (\textit{i.e.}, within $10^\circ$ margin).
Also, our method seems to be robust to a moderate amount of outliers, probably due to the inherent robustness of the $L_1$ optimization.

Additionally, to check if our estimates converged to a near-optimal solution, we compare them to the solution we would have got if we had used the ground truth as the initial seed.
We obtain this solution using the same method described in Section \ref{sec:calibration}, with the difference that we set $\mathbf{R}_\mathrm{est}$ to true $\widetilde{\mathbf{R}}_{mc}$ and $\theta_\mathrm{max}$ to $1^\circ$ in Step 1 and skip Step 6.
The result is shown in Fig. \ref{fig:calib_gt_init}.
We see that the angular differences between our estimates and the ground-truth-initialized ones are never above $0.04^\circ$.
This demonstrates that our calibration algorithm has a reliable convergence behavior.

\section{Discussion}
One limitation of the DTE is that it requires the knowledge of the camera-to-marker rotation $\widetilde{\mathbf{R}}_{mc}$ when obtaining the ground-truth camera orientations using \eqref{eq:DTE_Rmc}.
Although our calibration method in Section \ref{sec:calibration} works quite well under a reasonable amount of noise and outliers, the fact that it requires an additional data inquisition step may be considered cumbersome, especially when compared to the relatively simple process of computing the ATE.
Worse yet, if the ground-truth data had already been collected and if it does not contain the markers' orientations at all, then it would be simply impossible to compute the DTE.

That said, it would be an overstatement to say that this limitation alone significantly hinders the widespread adoption of the DTE.
First, there are cases where the ground-truth camera orientations $\widetilde{\mathbf{R}}_{gc,i}$ in \eqref{eq:DTE_rotation2} are directly available.
Examples include synthetic datasets (\textit{e.g.}, ICL-NUIM \cite{icl_nuim}, TartanAir \cite{tartan_air}) and real-world datasets where pseudo ground truth data is obtained by a structure-from-motion system\footnote{We refer to \cite{brachmann_2021_iccv} for an excellent discussion of this approach in visual relocalization.} (\textit{e.g.}, 1DSfM \cite{wilson_2014_eccv}, MVS \cite{zhou_2018_deeptam}). 
In such cases, $\widetilde{\mathbf{R}}_{mc}$ is simply the identity matrix.
Second, some of the most popular public datasets do provide the information about the relative rotation between the camera and the markers (or some sort of trackers with the same role). 
Examples include the TUM RGB-D \cite{sturm_2012_benchmark}, KITTI \cite{kitti} and EuRoC MAV \cite{euroc_mav} dataset.
Lastly, even if a public dataset provides only the markers' orientations without their relation to the camera's, it is still possible to estimate the camera-to-marker rotation using our calibration method in Section \ref{sec:calibration}, given that the cameras are sufficiently rotated with a varying axis and the estimated trajectory is reasonably accurate.
To ensure the latter, one could check whether or not the ATE is small enough. 
If it is too large, an accurate part of the trajectory can be extracted manually and used for calibration. 

Another limitation of the DTE is that it depends on the parameter $k$ we manually set in \eqref{eq:winsorized_error}.
The appropriate value for $k$ may differ according to the characteristics of each dataset.
Automatic tuning of $k$ is left for future work.
Also, the proposed method is not suitable when the outlier ratio is larger than 50\%.
In the future, we plan to explore other approaches using a more robust alignment method than the $L_1$ optimization (\textit{e.g.,} RANSAC \cite{ransac}).

\section{Conclusion}
In this work, we proposed the DTE, a novel metric for evaluating the trajectory estimation accuracy.
Unlike the ATE whose sensitivity quickly deteriorates in the presence of a few outliers, the DTE can robustly capture the varying accuracy as the inlier trajectory error or the number of outliers varies.
This is made possible by aligning the estimated trajectory to the ground truth using a robust method.
The key difference from the ATE is that we compute each of the translation, rotation and scale involved in the trajectory transformation using the geometric, geodesic and arithmetic median, respectively.
Furthermore, we winsorize and normalize the trajectory errors to limit the influence of unbounded outliers, while preventing misinterpretations of the results on small-scale datasets.
We also take the average of the mean and the RMS error to compute the DTE, which makes it behave favorably in terms of sensitivity to the inlier trajectory error, as well as the number of outliers.
Additionally, we proposed the DRE, a rotation-only metric that is based on the similar idea and has similar advantages to the DTE.
Lastly, we developed a simple yet effective algorithm for calibrating the camera-to-marker rotation, which is necessary for the computation of our metrics. 
In our calibration method, degeneracy was shown to occur when the relative camera rotations have the same axis of rotation (up to a sign). 
We evaluated our metrics and the calibration method through extensive simulations, demonstrating their effectiveness.

\section*{Appendix}
First, it is useful to know the following identity \cite{eade_2017_lie}:
\begin{equation}
    \label{eq:adjoint}
    \mathbf{R}\mathrm{Exp}(\mathbf{v})\mathbf{R}^\top = \mathrm{Exp}\left(\mathbf{R}\mathbf{v}\right).
\end{equation}
This means that for any rotation matrices $\mathbf{R}_1$ and $\mathbf{R}_2$, 
\begin{equation}
\label{eq:same_angle}
    \angle\left(\mathbf{R}_1\mathbf{R}_2\mathbf{R}_1^\top\right) = \angle\left(\mathbf{R}_2\right),
\end{equation}
where $\angle(\cdot)$ denotes the angle of the rotation.
When orientations are related by rotations around the same axis, they can be parameterized in two different ways using \eqref{eq:adjoint}:
\begin{equation}
\label{eq:same_axis_rotation}
    \mathbf{R}_i = \mathrm{Exp}(\theta_i\widehat{\mathbf{v}})\mathbf{R}_1 = \mathbf{R}_1\mathrm{Exp}(\theta_i\widehat{\mathbf{w}}) \quad \text{for all } i,
\end{equation}
where $\theta_i$ is the signed angle of rotation between the first and the $i$th orientation, and $\widehat{\mathbf{v}}$ and $\widehat{\mathbf{w}}$ are the axes of rotation that are related by $\widehat{\mathbf{v}}=\mathbf{R}_1\widehat{\mathbf{w}}$.

Next, we prove the following proposition:
When solving \eqref{eq:calibration}, degeneracy occurs if all cameras have the same fixed orientation, or if the axes of their relative rotations are all the same.
Since the former case is a special instance of the latter case (\textit{i.e.}, when the rotation angle is 0), we focus on the latter here.
First, we rewrite the summand in \eqref{eq:calibration} as
\begin{equation}
\label{eq:proof_si1}
    s_i = \angle\left(\widetilde{\mathbf{R}}_{gm,i}\widetilde{\mathbf{R}}_{mc}\mathbf{R}_{ec,i}^\top\mathbf{R}_{\text{align}}^\top\right).
\end{equation}
If the ground-truth orientations $\widetilde{\mathbf{R}}_{gm,i}$ are related to each other by rotations around the same axis (say $\widehat{\mathbf{v}}$), we have
\begin{equation}
\label{eq:proof_Rgmi}
    \widetilde{\mathbf{R}}_{gm,i}\stackrel{\eqref{eq:same_axis_rotation}}{=}\mathrm{Exp}(\theta_i\widehat{\mathbf{v}})\widetilde{\mathbf{R}}_{gm,1}.
\end{equation}
Now, consider the following update:
\begin{align}
    \left(\widetilde{\mathbf{R}}_{mc}\right)_\text{new}&\gets\widetilde{\mathbf{R}}_{gm,1}^\top\mathrm{Exp}(a\widehat{v})\widetilde{\mathbf{R}}_{gm,1}\widehat{\mathbf{R}}_{mc}, \label{eq:proof_Rmc}\\
    \left(\mathbf{R}_{\text{align}}\right)_\text{new}&\gets \mathrm{Exp}(a\widehat{v})\mathbf{R}_{\text{align}}, \label{eq:proof_Ralign}
\end{align}
where $a$ is an arbitrary angle.
Substituting \eqref{eq:proof_Rgmi}, \eqref{eq:proof_Rmc} and \eqref{eq:proof_Ralign} into \eqref{eq:proof_si1} and using the fact that the rotations are commutative if they share the same axis, we get
\begin{equation}
\label{eq:proof_si2}
    s_i = \angle\left(\mathrm{Exp}(a\widehat{\mathbf{v}})\widetilde{\mathbf{R}}_{gm,i}\widetilde{\mathbf{R}}_{mc}\mathbf{R}_{ec,i}^\top\mathbf{R}_{\text{align}}^\top\mathrm{Exp}(a\widehat{\mathbf{v}})^\top\right).
\end{equation}
According to \eqref{eq:same_angle}, the angle given by \eqref{eq:proof_si2} must be the same as that given by \eqref{eq:proof_si1}. 
Thus, by varying $a$ in \eqref{eq:proof_Rmc} and \eqref{eq:proof_Ralign}, we get infinitely many solutions for $\widetilde{\mathbf{R}}_{mc}$ and $\mathbf{R}_{\text{align}}$ that lead to the exact same cost in \eqref{eq:calibration}, resulting in degeneracy.

A similar argument can be made when the estimated orientations $\mathbf{R}_{ec,i}$ are related to one another by rotations around the same axis (say $\widehat{\mathbf{w}}$).
In this case, it is convenient to parameterize $\mathbf{R}_{ec,i}$ as follows:
\begin{equation}
\label{eq:proof_Reci}
    \mathbf{R}_{ec,i}\stackrel{\eqref{eq:same_axis_rotation}}{=} \mathbf{R}_{ec,1}\mathrm{Exp}(\theta_i\widehat{\mathbf{w}}).
\end{equation}
Now, consider the following update:
\begin{align}
    \left(\widetilde{\mathbf{R}}_{mc}\right)_\text{new}&\gets\widehat{\mathbf{R}}_{mc}\mathrm{Exp}(a\widehat{\mathbf{w}}), \label{eq:proof_Rmc2}\\
    \left(\mathbf{R}_{\text{align}}\right)_\text{new}&\gets \mathbf{R}_{\text{align}}\mathbf{R}_{ec,1}\mathrm{Exp}(a\widehat{\mathbf{w}})\mathbf{R}_{ec,1}^\top, \label{eq:proof_Ralign2}
\end{align}
where $a$ is an arbitrary angle.
Substituting \eqref{eq:proof_Reci}, \eqref{eq:proof_Rmc2} and \eqref{eq:proof_Ralign2} into \eqref{eq:proof_si1} and again using the fact that the rotations are commutative if they share the same axis, we end up with the same expression as \eqref{eq:proof_si1}.
Thus, by varying $a$ in \eqref{eq:proof_Rmc2} and \eqref{eq:proof_Ralign2}, we get infinitely many solutions that lead to the exact same cost in \eqref{eq:calibration}, resulting in degeneracy.

\addtolength{\textheight}{-12cm}   





\balance

\end{document}